\pdfoutput=1
\RequirePackage[T1]{fontenc}
\documentclass[conference,letterpaper]{IEEEtran}
\IEEEoverridecommandlockouts

\usepackage{cite}
\usepackage{amsmath,amssymb,amsfonts}
\usepackage{graphicx}
\usepackage{textcomp}
\usepackage{xcolor}
\usepackage[hidelinks,pdfversion=1.4]{hyperref}
\hypersetup{
  pdftitle={CenSynCMB: Centre Maps and Physics-Guided Synthesis for Microbleed Detection},
  pdfauthor={Lucas He et al.},
  pdfsubject={2026 IEEE International Conference on Bioinformatics and Biomedicine (BIBM)},
  pdfkeywords={cerebral microbleeds, MRI, lesion detection, synthetic data}
}

\usepackage{eso-pic}
\AddToShipoutPictureBG*{%
  \AtPageLowerLeft{%
    \put(\LenToUnit{0.5\paperwidth},\LenToUnit{0.32in}){%
      \makebox[0pt][c]{\parbox[b]{\textwidth}{\centering\scriptsize
        Accepted at the 2026 IEEE International Conference on Bioinformatics and Biomedicine (BIBM).\\
        \copyright~2026 IEEE. Personal use of this material is permitted. Permission from IEEE must be obtained for all other uses, in any current or future media, including reprinting/republishing this material for advertising or promotional purposes, creating new collective works, for resale or redistribution to servers or lists, or reuse of any copyrighted component of this work in other works.}}}}}
\def\BibTeX{{\rm B\kern-.05em{\sc i\kern-.025em b}\kern-.08em
    T\kern-.1667em\lower.7ex\hbox{E}\kern-.125emX}}

\begin{document}

\title{CenSynCMB: Centre Maps and Physics-Guided Synthesis for Microbleed Detection}

\author{
\IEEEauthorblockN{\begin{tabular}{@{}c@{}}
Lucas He\textsuperscript{1,2}\thanks{Corresponding author: Lucas.he.23@ucl.ac.uk.},
Hanyuan Zhang\textsuperscript{1},
Krinos Li\textsuperscript{3},
Adama Fatima Saccoh\textsuperscript{4} \\
Silvia Ingala\textsuperscript{5},
Rafael Rehwald\textsuperscript{6},
Marleen de Bruijne\textsuperscript{7},
Frederik Barkhof\textsuperscript{1,8,9} \\
Rhodri Davies\textsuperscript{2,10,\textdagger}\thanks{Joint senior authors.},
Carole H. Sudre\textsuperscript{1,2,11,\textdagger}
\end{tabular}}
\IEEEauthorblockA{\begin{minipage}{0.99\textwidth}
\centering\scriptsize
\textsuperscript{1}Hawkes Institute, University College London, UK\\
\textsuperscript{2}Unit for Lifelong Health and Aging, University College London, UK\\
\textsuperscript{3}Bioengineering Department and Imperial-X, Imperial College London, UK\\
\textsuperscript{4}Institute of Cardiovascular Science, University College London, UK\\
\textsuperscript{5}Department of Diagnostic Radiology, Copenhagen University Hospital, Denmark\\
\textsuperscript{6}Neuroradiological Academic Unit, UCL Queen Square Institute of Neurology, London, UK\\
\textsuperscript{7}Department of Radiology and Nuclear Medicine, Erasmus MC, The Netherlands\\
\textsuperscript{8}Department of Radiology \& Nuclear Medicine, Amsterdam UMC, Vrije Universiteit, The Netherlands\\
\textsuperscript{9}Queen Square Institute of Neurology, University College London, UK\\
\textsuperscript{10}Barts Heart Centre, St Bartholomew's Hospital, London, UK\\
\textsuperscript{11}Department of Biomedical Computing - School of Biomedical Engineering and Imaging Sciences,\\
King's College London, UK
\end{minipage}}
}

\maketitle

\begin{abstract}
Cerebral microbleeds (CMBs) are MRI markers of small vessel disease and the microbleed component of amyloid-related imaging abnormalities (ARIA-H), but their small size, sparsity, and similarity to vessels, calcification-like foci, and artefacts make automated detection difficult. We propose CenSynCMB, a centre-guided and mimic-aware framework combining a 3D Attention U-Net, auxiliary centre-map supervision, false-negative-driven reweighting, and fold-wise physics-guided synthesis of positive CMBs and labelled hard negatives. Synthetic data expose the detector to compact lesions and common mimics without validation or test leakage. On VALDO Task~2, CenSynCMB achieved the best local-comparison lesion-level F1 (74.3$\pm$8.8\%, $p=0.020$); on external AIBL SWI, it achieved the highest local-comparison recall (88.5$\pm$6.9\%, $p=0.0058$) and F1 (65.0$\pm$6.9\%, $p=0.0016$). Together, these results support scalable CMB candidate extraction in large, unlabelled MRI cohorts, while highlighting cohort-specific calibration as the next step toward reliable burden estimation.
We have open-sourced the inference implementation and released pretrained model weights at \url{https://github.com/LucasHe17/CenSynCMB_Microbleed}.
\end{abstract}


\section{Introduction}

\subsection{Clinical Motivation}
Large studies assessing cerebral small vessel disease (CSVD) increasingly require more than visual summaries of lesion burden. CSVD is characterised on MRI by markers such as white matter hyperintensities (WMH), lacunes, enlarged perivascular spaces (EPVS), and cerebral microbleeds (CMBs)~\cite{strive}. Among these markers, CMBs appear as small rounded hypointense foci on susceptibility-sensitive MRI such as T2$^*$ or susceptibility-weighted imaging (SWI), with diagnostic criteria commonly restricting them to lesions below 10\,mm in diameter~\cite{greenberg2009,bombs}. They carry information about vascular brain injury, ageing, cognitive decline, and dementia-related pathology~\cite{greenberg2009}. Their anatomical distribution is clinically informative: lobar CMBs are more often associated with cerebral amyloid angiopathy, whereas deep CMBs are more often linked to hypertensive arteriopathy~\cite{greenberg2009}. Accurate CMB quantification is therefore important not only for image interpretation and cohort studies, but also for monitoring amyloid-related imaging abnormalities with haemorrhage or haemosiderin deposition (ARIA-H) during anti-amyloid monoclonal antibody therapy~\cite{aria2024}.

\begin{figure*}[t]
\centering
\includegraphics[width=1.0\textwidth]{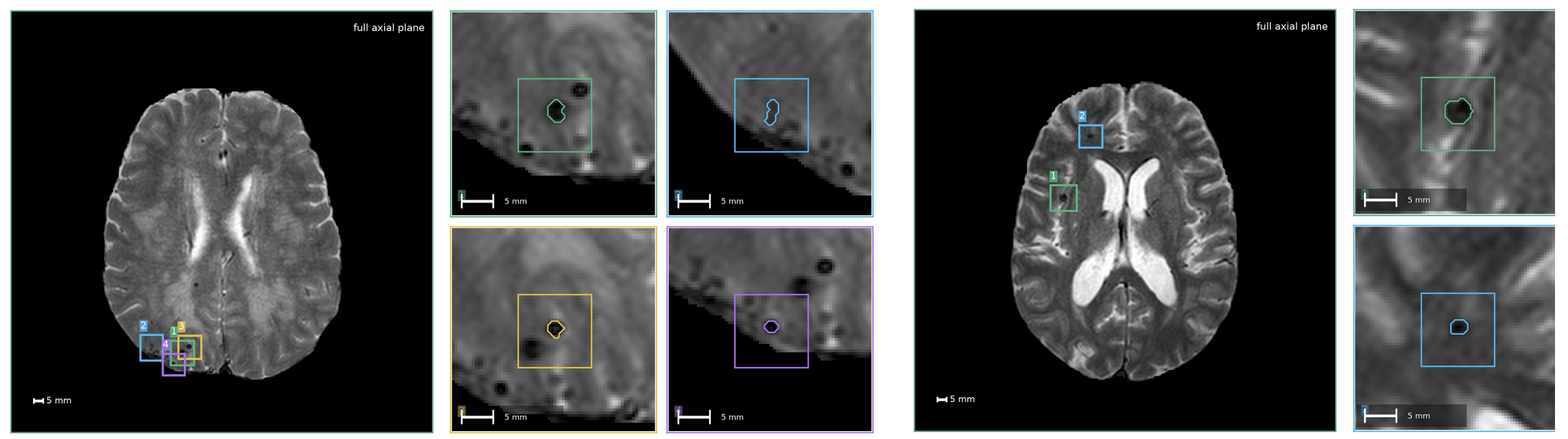}
\caption{Representative appearance of cerebral microbleeds (CMBs) on T2*-weighted gradient-echo MRI (T2*-GRE, VALDO), shown for two subjects. Coloured boxes and enlarged crops show individual CMBs as small rounded hypointense foci with reference contours. Scale bars: 5\,mm.}
\label{fig:intro-cmb}
\end{figure*}

Despite this need, many CSVD studies still rely on visual assessment, including binary presence ratings, lesion counts, or structured scales such as the Microbleed Anatomical Rating Scale (MARS) and Brain Observer MicroBleed Scale (BOMBS)~\cite{mars,bombs}. These scales improve standardisation, but they still require trained readers and remain time-consuming for large datasets. The Vascular Lesions Detection and Segmentation (VALDO) challenge found that visual assessment is affected by inter- and intra-rater variability, making robust automated quantification important~\cite{valdo}. Fig.~\ref{fig:intro-cmb} illustrates why the task remains challenging even before automation: CMBs are small, spatially sparse, and reliably visible only on susceptibility-sensitive sequences under focal, slice-by-slice inspection.

\subsection{Detection Challenges}
Automated CMB detection is difficult because the targets are compact, sparse, and imaged with anisotropic susceptibility-sensitive acquisitions in which slice thickness can approach or exceed the lesion diameter. Their rounded morphology makes the lesion centre a natural localisation proxy, but the surrounding appearance is ambiguous: small vessels in cross-section, calcification-like foci, susceptibility artefacts, and noise can create small hypointense signals that resemble true microbleeds~\cite{greenberg2009}. Co-registered T1- and T2-weighted images can help disambiguate some mimics through cross-contrast appearance, while phase imaging or quantitative susceptibility mapping (QSM) may be needed when calcium and blood products must be separated by susceptibility sign. Acquisition protocol also changes apparent lesion size and burden, so models trained on T2$^*$-GRE may not transfer directly to SWI~\cite{greenberg2009}.

Benchmark evidence suggests that automated CMB detection is feasible, but not yet settled. Task~2 of VALDO (VAscular Lesion Detection and Segmentation) at MICCAI, showed promising performance while also highlighting large variability across methods and cases and a gap between population-level utility and reliable individual-level deployment~\cite{valdo}. The GRE-to-SWI shift is therefore not only a deployment concern, but also an evaluation requirement, as recent GRE/SWI CMB frameworks also treat cross-sequence robustness as a central goal~\cite{hassanzadeh2025}. In this work we train on VALDO T2$^*$-GRE data and test externally on the AIBL SWI cohort. Public pretrained models are reported as reference anchors when their supervision may overlap with benchmark or external datasets, but they are kept distinct from leakage-free replicated models.

Two technical bottlenecks are especially relevant for CMB detection. The first is a mismatch between the training objective and the evaluation target: compact CMBs are scored as object centroids rather than as large anatomical segmentations, so segmentation backbones trained only through voxel-level overlap can under-emphasise centroid accuracy. The second is the scarcity of hard negatives. Sensitivity-oriented training can increase false positives unless it is paired with realistic negative examples, yet real datasets contain limited examples of the mimic patterns that often drive false positives. A model that has seen only positive lesions and generic background may therefore learn the appearance of CMBs without learning enough about nearby mimics.

\subsection{Contributions}
We address these bottlenecks with CenSynCMB, a centre-guided CMB detector trained with physics-guided hard-negative synthesis. The detector uses an Attention U-Net backbone, an auxiliary centre-map head, and a Tversky objective with false-negative-driven crop reweighting to favour small-lesion recovery. The synthesis module then supplies labelled vessel-like and calcification-like mimics, providing the negative evidence needed to preserve precision in a single-stage detector.

\begin{itemize}
\item We propose CenSynCMB, a centre-guided CMB detector that combines centre-map supervision, focal centre-map regression, uncertainty-weighted multitask learning, and false-negative-driven reweighting within a simple probability-map-to-centroid inference path.
\item We introduce a fold-wise physics-guided synthesis module that renders synthetic positive CMBs and labelled vessel-like and calcification-like hard negatives into real training backgrounds, providing supervision for both rare lesions and clinically plausible mimics without using validation or test subjects.
\end{itemize}

We evaluate CenSynCMB on VALDO Task~2 and on AIBL as an external SWI test cohort. The experiments include lesion-level detection, patient-level burden regression, comparison with current methods, and ablations isolating the effects of centre-map supervision, false-negative-driven reweighting, synthetic positive CMBs, and synthetic hard negatives.

\section{Related Work}

Although many cohort studies still summarise CSVD burden with visual ratings, the methodological literature has increasingly moved toward lesion-level detection and quantitative readouts. The VALDO challenge established a shared benchmark for EPVS, CMBs, and lacunes, showing both the promise of automated tools at population level and their inconsistency across individual cases~\cite{valdo}. Work on lacunes has faced a related combination of small lesion size, sparse labels, and radiological mimics; for example, SWIN-DS introduced deeply supervised transformer features with geometric guidance for robust lacune detection~\cite{he2026swinds}. Joint CSVD-marker modelling has also been explored through cross-task attention and anatomically informed calibration for lacunes and enlarged perivascular spaces~\cite{he2026unified}. These studies show that small CSVD markers benefit from lesion-specific supervision and false-positive control, but CMBs require different imaging cues and mimic modelling because they are susceptibility-driven hypointense foci rather than CSF-like cavities.

CMB detection methods span general-purpose segmentation architectures and task-specific designs. General 3D segmentation backbones, including nnU-Net-style networks, Attention U-Net, SwinUNETR, and foundation-style segmentation models, provide strong reusable architectures but are usually trained through dense overlap losses~\cite{nnunet,attentionunet,swinunetr,vista3d}. CMB-specific systems, including VALDO challenge methods, MixMicrobleedNet, Al-Masni-style two-stage detectors, Kim-style proposal networks, FRST-style detectors, and SHIVA-CMB, generally gain precision through cascaded false-positive reduction or task-specific priors, but do not couple explicit centroid supervision with labelled mimic examples in a single-stage design~\cite{cmbseg,mixmicrobleednet,almasni,kim2025,sundaresan,shivacmb}. Recent work has also targeted GRE/SWI robustness with CNN--YOLO pipelines~\cite{hassanzadeh2025} and multicentre CMB quantification for ARIA-H monitoring~\cite{low2026}. These methods demonstrate that CMB detection is feasible, but they differ in training data, availability of public weights, sequence assumptions, and potential overlap with benchmark cohorts. Therefore, direct comparison requires a common centroid-matching protocol and a clear distinction between locally replicated models and open-source pretrained references.

Complementary work has incorporated anatomical and subject-level priors into CMB segmentation. Kwon et al. introduced auxiliary segmentation of lobar, deep supratentorial, and infratentorial regions to guide CMB localisation and false-positive reduction~\cite{kwon2024anatomy}, and explored masked-image pretraining of UNETR++ to address limited labelled data~\cite{kwon2024unetr}. Their BP-nnUNet integrates estimated blood-pressure features with lobar/deep microbleed prompts~\cite{kwon2025bp}, while AgeCMBNet uses estimated age to select age-specific decoder branches with a shared encoder~\cite{kwon2026age}. These studies demonstrate the value of anatomical context, pretrained representations, and subject-level information. CenSynCMB focuses on complementary lesion-level signals: explicit centre-map supervision and controlled synthesis of positive lesions and labelled mimics.

Synthetic data has been widely used to mitigate annotation scarcity in medical imaging, but its use in CMB detection remains limited, especially for labelled mimic phenotypes. Conventional augmentation changes the appearance of existing examples without introducing new lesion-mimic configurations, while learned image generators can be difficult to control and may not follow plausible imaging mechanisms. For CMBs, a useful synthetic route should preserve real brain background, respect gyrus-sulcus boundaries, and provide supervision for both positive lesions and their mimics. The present work therefore focuses on a physics-guided rendering of synthetic positive CMBs together with labelled vessel-like and calcification-like hard negatives, complementing the centre-guided detector with explicit mimic supervision.

\section{Methods}

\subsection{Pipeline Overview}
\begin{figure*}[t]
\centering
\includegraphics[width=1.0\textwidth]{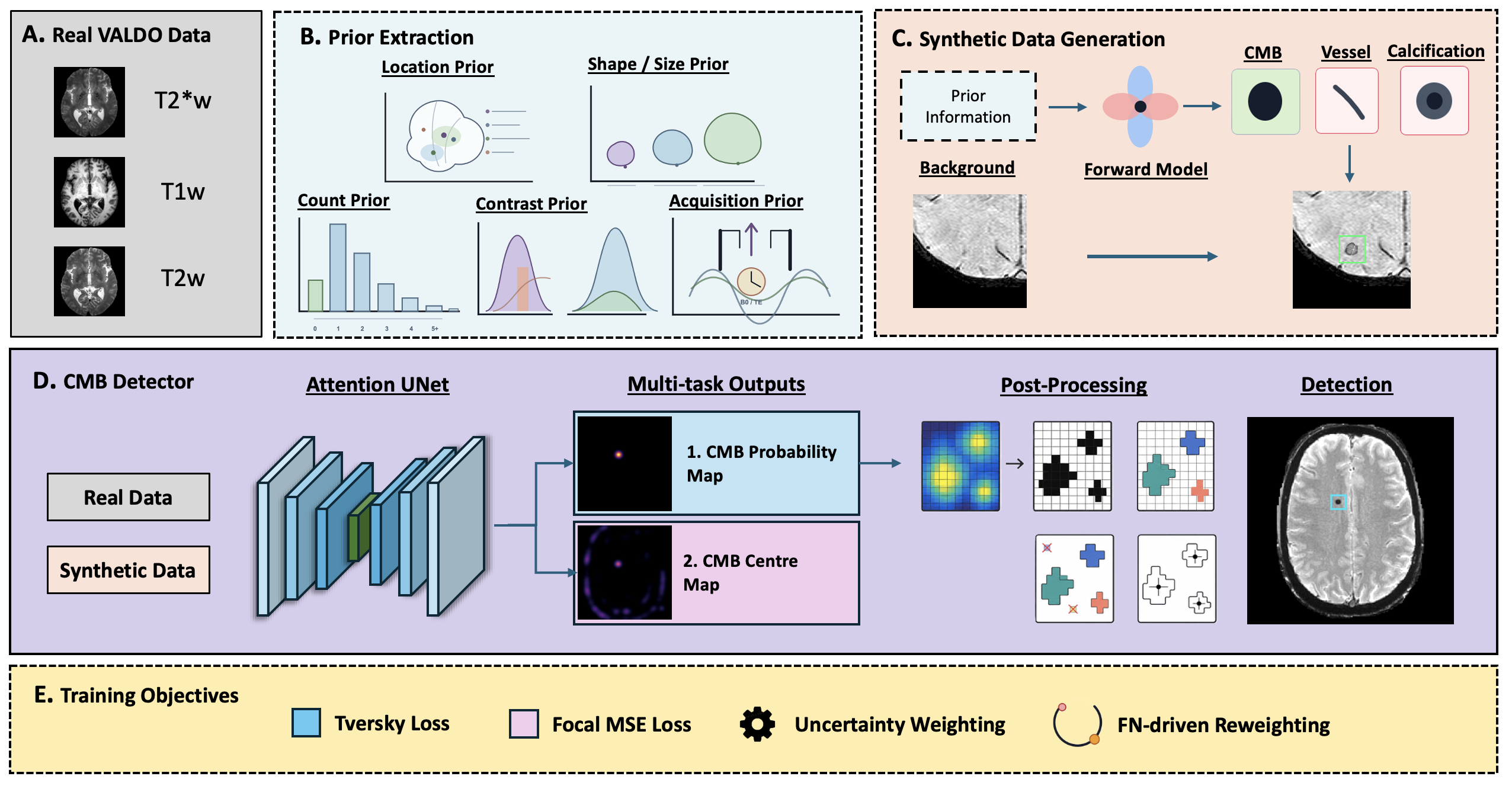}
\caption{Overview of the centre-guided, synthesis-augmented CMB detection framework. (A) Real VALDO data provide T2S-space co-registered T2$^*$-weighted (T2$^*$w), T1-weighted (T1w), and T2-weighted (T2w) MRI volumes. (B) Training-fold statistics are used to estimate lesion location, count, shape/size, and acquisition priors. (C) Physics-guided synthesis inserts positive cerebral microbleeds (CMBs) and labelled vessel-like and calcification-like hard negatives into real image backgrounds. (D) Real and synthetic subjects train an Attention U-Net with a CMB probability map and an auxiliary CMB centre-map head; detections are obtained by connected-component post-processing and centroid extraction. (E) The objective combines Tversky loss, focal mean-squared error (MSE) centre-map loss, homoscedastic uncertainty weighting, and false-negative (FN)-driven reweighting.}
\label{fig:cmb-overview}
\end{figure*}

CenSynCMB combines a centre-guided CMB detector with physics-guided synthetic training data (Fig.~\ref{fig:cmb-overview}). We use the VALDO-released T1, T2, CMB mask, and T2$^*$-GRE volumes in T2S (T2$^*$-GRE) space, so the detector receives co-registered multi-contrast inputs on the susceptibility-sensitive reference grid and produces CMB centroids directly from its probability map. This design keeps inference simple while shifting the training signal toward small-lesion localisation rather than voxel overlap alone. During network preprocessing, images and label maps are resampled jointly to 1\,mm isotropic spacing using continuous interpolation for images and nearest-neighbour interpolation for masks.

Physics-guided synthesis is used only during training. It augments the detector training pool with fold-wise synthetic positive CMBs and labelled hard negatives, leaving validation and test inference entirely on real images.

\subsection{Centre-Guided CMB Detector}
\label{subsec:detector}
The detector outputs a set of microbleed centroids from a dense probability map. We use a 3D Attention U-Net~\cite{attentionunet} as the backbone because its attention gates can dampen broad normal-anatomy responses while preserving local convolutional features around small hypointense lesions. This inductive bias is well matched to compact-lesion detection, where the target is sparse and annotated examples are limited.

Given co-registered T1, T2, and T2$^*$-GRE volumes, the network predicts a two-channel CMB probability map and an auxiliary centre map. The probability map is converted into detections by fixed post-processing: thresholding at 0.50, 26-connected-component grouping, removal of components smaller than five voxels, and reduction of each remaining component to its centroid. These values were fixed for all reported evaluations rather than tuned on external data. This defines the final detection representation used in evaluation; the auxiliary centre map is not queried at inference time.

The centre-map head supplies a centre-localisation signal that is closer to the detection target than voxel overlap alone. For each connected reference lesion with centroid $c_k$, we define a Gaussian target
\begin{equation}
C(x)=\max_k \exp\!\left(-\frac{\lVert x-c_k\rVert^2}{2\sigma_c^2}\right),
\end{equation}
where $\sigma_c=2$ voxels. This head trains the shared representation to localise lesion centres, but it does not replace the probability-map proposal path at inference.

\subsection{Training Objective}
\label{subsec:training-objective}
The training objective combines a Tversky segmentation term, a focal centre-map regression term, and uncertainty-weighted multi-task balancing:
\begin{equation}
\begin{aligned}
\mathcal L_{\mathrm{seg}} &=
1-\frac{\mathrm{TP}+\epsilon}
{\mathrm{TP}+\alpha\,\mathrm{FP}+\beta\,\mathrm{FN}+\epsilon},\\
\mathcal L_{\mathrm{ctr}} &=
\frac{1}{|\Omega|}\sum_{x\in\Omega}(C_x+\epsilon_c)^{\gamma_c} \\
&\quad \times \left(\sigma(\hat C_x)-C_x\right)^2,\\
\mathcal L_{\mathrm{base}} &=
\mathcal L_{\mathrm{seg}}+\exp(-s_{\mathrm{ctr}})\mathcal L_{\mathrm{ctr}}+s_{\mathrm{ctr}},
\end{aligned}
\end{equation}
The centre-map logit is $\hat C_x$, $\sigma(\cdot)$ is the sigmoid function, and $s_{\mathrm{ctr}}$ is the learnable log variance for homoscedastic uncertainty weighting~\cite{kendall2018}. We set $\alpha=0.1$ and $\beta=0.9$, placing higher cost on false negatives than false positives, and use $\gamma_c=2.5$ and $\epsilon_c=10^{-3}$ for the focal centre-map term.

Finally, we apply false-negative-driven crop reweighting. For lesion $k$ with voxel set $S_k$, size $|S_k|$, and current in-lesion response $r_k=\max_{x\in S_k}\hat p_x$, the crop weight and final loss are
\begin{equation}
\begin{aligned}
w_k &=
\left(\frac{\bar s}{|S_k|+\epsilon}\right)^{\gamma_s} \\
&\quad \times \left(1+\rho\,\mathrm{clip}(1-r_k,0,1)\right),\\
w_{\mathrm{crop}} &= \mathrm{clip}\!\left(\max_k w_k,w_{\min},w_{\max}\right),\\
\mathcal L_{\mathrm{crop}} &= w_{\mathrm{crop}}\mathcal L_{\mathrm{base}}.
\end{aligned}
\end{equation}
We use $\gamma_s=0.5$, $\rho=1.5$, and clip the crop weight to $[0.5,5.0]$. This targeted reweighting is more specific than generic hard-example mining: it jointly emphasises small lesions and lesions that the current detector is responding to weakly.

\subsection{Physics-Guided Synthesis}
The synthesis module augments detector training with controllable examples that are difficult to collect in sufficient numbers from the training set alone. Rather than training a generative model, we render physics-guided positive CMBs and labelled hard negatives directly into real training backgrounds. This keeps the surrounding brain anatomy, scanner texture, and intensity statistics real while changing only the local lesion or mimic region.

All synthesis is performed within each cross-validation fold. Positive-lesion priors are estimated only from connected CMB components in the training subjects of that fold: we store per-subject lesion counts, relative lesion centres, component voxel offsets, equivalent diameters, and local multi-contrast intensity changes, then resample from these empirical summaries during insertion. Host volumes are also sampled only from the same training split. Validation and test subjects are never used to fit priors, choose host images, tune synthetic appearance, or generate training examples. The synthetic data are therefore leakage-free with respect to the held-out fold.

For the T2$^*$-GRE channel, we approximate the local blooming pattern with a susceptibility-inspired forward model. A binary source mask is assigned a scalar susceptibility amplitude $\chi$ and transformed in k-space by multiplication with the standard dipole kernel~\cite{salomir2003}
\begin{equation}
D(k)=\frac{1}{3}-\frac{k_z^2}{\lVert k\rVert^2},
\end{equation}
equivalently convolution with the dipole kernel in real space, to obtain the local field perturbation. Under a static-dephasing approximation, spins inside a voxel acquire phase $\phi(x)=\gamma\,\Delta B_z(x)\,\mathrm{TE}$, and the synthesized magnitude attenuation is obtained from the modulus of the voxel-averaged complex signal. If reliable field-strength or echo-time metadata are unavailable, we sample 1.5--3.0\,T and 20--40\,ms for domain randomization. We treat this renderer as a susceptibility-inspired approximation for training augmentation, not as a full GRE or QSM simulator.

\begin{figure}[t]
\centering
\includegraphics[width=1.0\columnwidth]{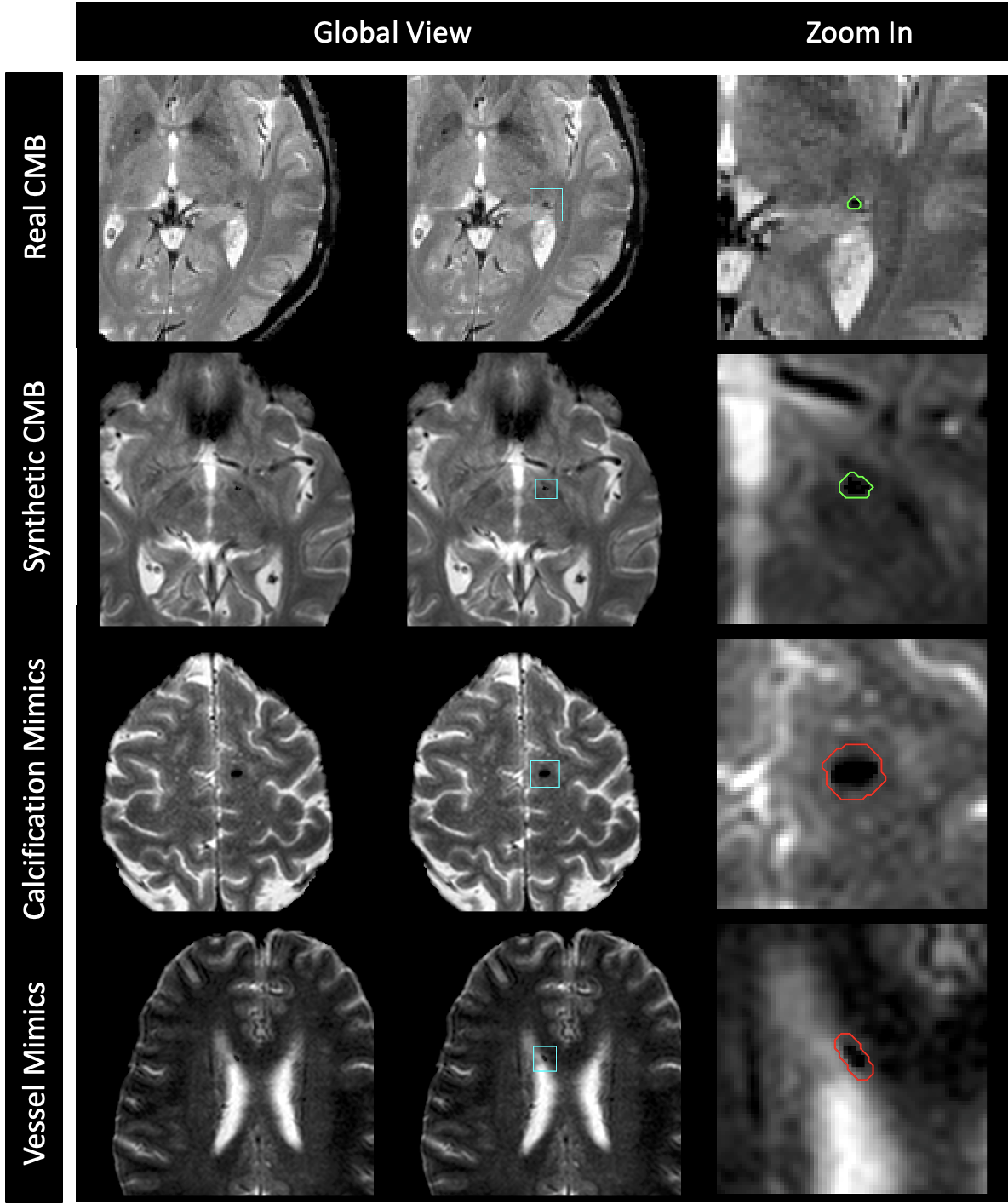}
\caption{Qualitative examples of real and synthesized CMB-related appearances. Rows show real CMBs, synthetic positive CMBs, calcification-like mimics, and vessel-like mimics. Columns show the full T2$^*$-weighted slice, the target-box overlay, and an enlarged 64$\times$64 crop with the mask boundary. Green contours denote CMB masks, while red contours denote labelled hard-negative mimic masks.}
\label{fig:synthetic-examples}
\end{figure}

Synthetic positive CMBs are inserted as compact susceptibility sources sampled from these training-lesion priors, and their masks are saved to generate the same segmentation and centre-map targets as real lesions. Hard-negative priors are constructed separately around common false-positive phenotypes. Calcification-like negatives use compact low-signal sources with non-CMB multi-contrast patterns; they are not distinguished by susceptibility sign because a magnitude-only static-dephasing model is insensitive to a global sign flip of $\chi$, the same ambiguity that clinical reading resolves using SWI phase or QSM. Vessel-like negatives use elongated anisotropic masks to create tubular low-signal structures on T2$^*$-GRE. Fig.~\ref{fig:synthetic-examples} shows representative real and synthesized appearances, and Table~\ref{tab:data-synthesis-ablation} isolates the effects of synthetic positive CMBs and synthetic hard negatives during detector training.

\section{Experiments}

\subsection{Dataset}
We developed and selected all models on VALDO Task~2 and used AIBL only for external inference. VALDO Task~2 provided 72 subjects from three population cohorts SABRE, RSS, and ALFA, including 50 CMB-positive and 22 CMB-negative cases, with co-registered T2, T2$^*$-GRE, and T1 volumes in T2S (T2$^*$-GRE) space and voxel-level CMB masks~\cite{valdo}. Native T2$^*$ voxel spacing ranged from 0.45--1.00\,mm in-plane and 0.80--4.00\,mm through-plane before our 1\,mm isotropic preprocessing. The cohort was enriched for CMB-positive cases relative to general population prevalence, while lesion counts still varied across positive subjects; acquisition spanned 1.5\,T and 3\,T scanners from different vendors~\cite{valdo}.

The external cohort came from the Australian Imaging, Biomarkers and Lifestyle (AIBL) study of ageing through CSIRO collection~50304~\cite{momeni2021}. It introduced a deliberate GRE-to-SWI shift and contained 370 prepared SWI volumes, including 57 CMB-positive and 313 CMB-negative cases, acquired on a 3\,T Siemens Tim Trio scanner at 0.93$\times$0.93$\times$1.75\,mm with a 20\,ms echo time. AIBL was not used for training or model selection of any locally replicated model.

\subsection{Implementation Details}

\subsubsection{Software and Hardware}
All models were implemented in Python with PyTorch (v2.6.0) and MONAI (v1.5.0)~\cite{monai}. Training used NVIDIA GH200 GPUs with 80\,GB memory per device on the Isambard-AI National AI Research Resource, operated by the Bristol Centre for Supercomputing (BriCS)~\cite{isambardai}, with fixed per-fold random seeds.

\subsubsection{Detector Training}
The detector was a 3D Attention U-Net with co-registered T1, T2, and T2$^*$-GRE channels. Volumes were reoriented to RAS, resampled to 1\,mm isotropic spacing, scaled between the 0.5th and 99.5th intensity percentiles, and z-score normalized per channel over nonzero voxels. Training used six 96$^3$ patches, sampled at a 2:1 lesion-centred/background ratio with random flips and gamma augmentation. The network used encoder widths from 32 to 512, dropout 0.1, and the objective in Section~\ref{subsec:training-objective}. We trained for 200 epochs with AdamW (learning rate $5\times10^{-4}$, weight decay $10^{-5}$), mixed precision, LR reduction on validation-F1 plateaus, and best-F1 checkpointing. Inference used 128$^3$ sliding windows with 0.6 overlap.

\subsubsection{Synthetic Augmentation}
Synthesis-enabled runs used fold-specific offline synthetic manifests. Real and synthetic subjects were sampled at a 1:1 ratio and passed through the same preprocessing and training pipeline; validation, checkpoint selection, and all reported evaluations used real images only.

\subsection{Evaluation Protocol}
We carried out all training and validation on VALDO Task~2 with 5-fold cross-validation stratified by CMB presence, fixing the partitions with a single random seed. Within each fold the model was trained on 4 folds and assessed on the held-out fold, and every VALDO result was aggregated across the five held-out test folds so that each subject was scored exactly once. Following the VALDO Task~2 detection criterion, a predicted centroid after the detector post-processing in Section~\ref{subsec:detector} counted as a true positive when it lay within 5\,mm of a reference CMB centroid; assignments were resolved one-to-one by greedy nearest matching, leaving unmatched predictions as false positives and unmatched references as false negatives.

At the lesion level we reported Recall, Precision, F1, and false positives per subject, computed per subject and then macro-averaged over the pooled test subjects. At the patient level we assessed burden estimation by comparing predicted and reference lesion counts, reporting mean absolute error (MAE), root mean squared error (RMSE), and Pearson correlation. Every metric was accompanied by a 95\% confidence interval from a non-parametric bootstrap that resampled subjects with replacement over 500 iterations and took the 2.5th and 97.5th percentiles, summarised as half-widths in the tables.

\subsection{Statistical Analysis}
All statistical comparisons were paired and descriptive. Per-subject metrics were compared with a two-sided Wilcoxon signed-rank test. Pearson correlations were reported with paired subject-bootstrap confidence intervals, but were not used as confirmatory significance endpoints. We reported nominal $p$ values without multiplicity correction because the tests support pre-specified, correlated metrics rather than independent confirmatory hypotheses.

\begin{figure}[t]
\centering
\includegraphics[width=\columnwidth]{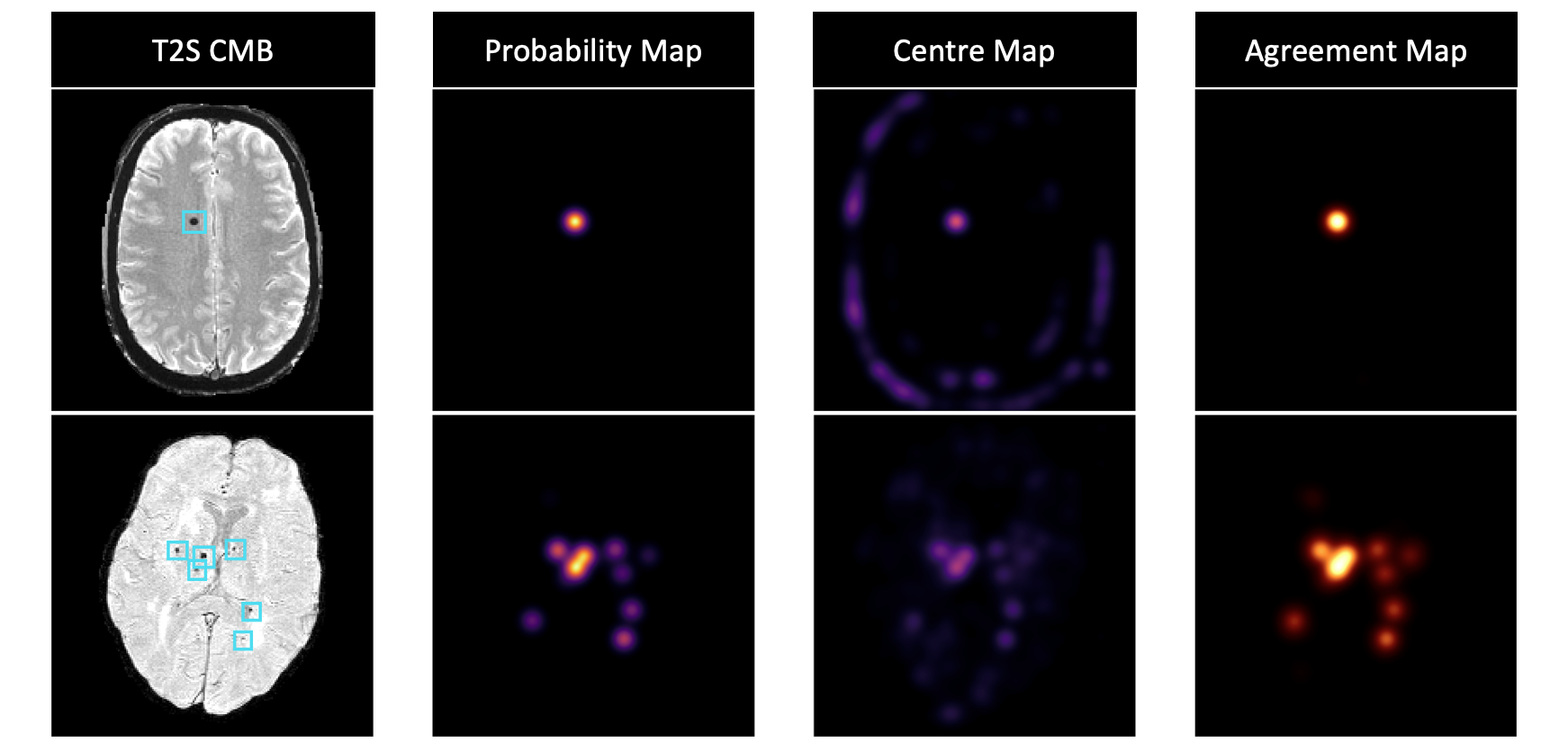}
\caption{Grad-CAM visualisation of the centre-guided detector outputs, showing the T2$^*$-weighted slice, probability-map head, centre-map head, and agreement map. The centre-map head provides a complementary localisation signal around compact CMB candidates.}
\label{fig:grad-cam}
\end{figure}

\subsection{Comparison Methods}
We tested our detector against three groups of methods under the same centroid-matching protocol, so any difference reflects the method and not the evaluation. Methods we reimplemented were retrained on each fold using our splits (R). Public pretrained models (L) serve only as contextual references, not fair baselines, because their released weights were not produced under our fold-wise protocol; SHIVA-CMB v2 also used AIBL material in its original development and evaluation. They are therefore excluded from bold/underline ranking and statistical comparator selection.

\textbf{Segmentation baselines.} We trained four general-purpose 3D segmentation networks inside our pipeline: DynUNet, a self-configuring nnU-Net-style backbone~\cite{nnunet}; Attention U-Net~\cite{attentionunet}, the architecture we also adopt for the centre-guided detector; SwinUNETR~\cite{swinunetr}, a Swin-transformer encoder with a convolutional decoder; and VISTA3D~\cite{vista3d}, a foundation-style promptable encoder-decoder adapted to microbleeds through a learned class prompt. Trained from scratch under identical settings, they form leakage-free internal references.

\textbf{VALDO challenge methods.} Zihao VALDO~\cite{cmbseg} is the multi-stage screening, discrimination, and segmentation framework that placed third in VALDO Task~2; we adapted its public code and retrained it per fold (R), using none of its released weights. MixMicrobleedNet~\cite{mixmicrobleednet} is also a VALDO Task~2 submission, but we used its public pretrained weights rather than retraining it within our cross-validation splits; it is therefore shown as a contextual pretrained reference (L), not as a ranked local comparator.

\textbf{Recent CMB detectors.} Three recent detectors were reimplemented under our protocol (R): an Al-Masni-style detector pairing a regional YOLO stage with a 3D-CNN false-positive classifier~\cite{almasni}; a Kim-style detector built on a U-Net with a region-proposal network, feature fusion, and hard-sample prototype learning~\cite{kim2025}; and an FRST-style detector that seeds candidates with a fast radial-symmetry transform on T2$^*$-GRE and refines them with a CNN/V-Net~\cite{sundaresan}.

\section{Results}

\subsection{Comparison with Current Methods}

\subsubsection{Lesion Level Detection}

Table~\ref{tab:lesion-level-comparison} summarises lesion-level detection on the VALDO development cohort and the AIBL external SWI cohort under the shared centroid-matching protocol. The comparison includes MONAI baselines, VALDO-family methods, current CMB detectors, pretrained reference models, and CenSynCMB. The central pattern is that CenSynCMB improved lesion-level F1 under both the in-domain VALDO setting and the harder GRE-to-SWI transfer setting, but the external gain came with a higher false-positive burden.

\begin{table*}[t]
\centering
\caption{Lesion-level comparison.}
\label{tab:lesion-level-comparison}
\scriptsize
\setlength{\tabcolsep}{2pt}
\renewcommand{\arraystretch}{1.12}
\resizebox{\textwidth}{!}{%
\begin{tabular}{llcccc@{\hspace{10pt}}cccc}
\hline
 &  & \multicolumn{4}{c}{VALDO (n=72)} & \multicolumn{4}{c}{AIBL (n=370)} \\
\cline{3-6}\cline{7-10}
Group & Method & Recall $\uparrow$ (\%) & Precision $\uparrow$ (\%) & F1 $\uparrow$ (\%) & FP/subject $\downarrow$ & Recall $\uparrow$ (\%) & Precision $\uparrow$ (\%) & F1 $\uparrow$ (\%) & FP/subject $\downarrow$ \\
\hline
Baseline & DynUNet~\cite{nnunet} & 69.7 $\pm$ 11.1 & 61.9 $\pm$ 10.8 & 61.3 $\pm$ 10.3 & 1.3 $\pm$ 0.5 & 53.5 $\pm$ 6.8 & \underline{53.1 $\pm$ 6.4} & \underline{49.4 $\pm$ 7.4} & \textbf{0.9 $\pm$ 0.3} \\
 & Attention U-Net~\cite{attentionunet} & 69.7 $\pm$ 10.3 & \underline{68.6 $\pm$ 10.3} & \underline{65.8 $\pm$ 10.2} & 1.4 $\pm$ 0.9 & 62.2 $\pm$ 3.4 & 25.1 $\pm$ 4.4 & 29.2 $\pm$ 4.9 & 7.6 $\pm$ 1.0 \\
 & SwinUNETR~\cite{swinunetr} & 64.2 $\pm$ 10.5 & 54.2 $\pm$ 10.5 & 54.3 $\pm$ 10.4 & 2.7 $\pm$ 1.8 & 57.9 $\pm$ 1.4 & 14.9 $\pm$ 4.0 & 20.6 $\pm$ 3.8 & 11.0 $\pm$ 1.2 \\
 & VISTA3D~\cite{vista3d} & 64.0 $\pm$ 11.0 & 58.4 $\pm$ 10.5 & 56.4 $\pm$ 10.0 & 1.5 $\pm$ 0.6 & 66.2 $\pm$ 2.4 & 25.7 $\pm$ 4.4 & 30.6 $\pm$ 4.6 & 8.4 $\pm$ 1.2 \\
\hline
VALDO methods & Zihao VALDO~\cite{cmbseg}\textsuperscript{R} & \underline{70.8 $\pm$ 9.8} & 61.8 $\pm$ 10.2 & 61.1 $\pm$ 9.2 & 2.0 $\pm$ 1.4 & 63.5 $\pm$ 1.9 & 23.5 $\pm$ 4.1 & 29.2 $\pm$ 4.8 & 7.8 $\pm$ 1.1 \\
\hline
Current methods & Al-Masni-style detector~\cite{almasni}\textsuperscript{R} & 45.1 $\pm$ 12.5 & 36.9 $\pm$ 10.4 & 33.6 $\pm$ 8.8 & 2.4 $\pm$ 1.0 & 48.6 $\pm$ 1.2 & 16.9 $\pm$ 4.0 & 23.0 $\pm$ 4.2 & 7.3 $\pm$ 0.7 \\
 & Kim-style detector~\cite{kim2025}\textsuperscript{R} & 64.7 $\pm$ 11.3 & 65.2 $\pm$ 10.9 & 61.0 $\pm$ 10.2 & \underline{0.9 $\pm$ 0.4} & 53.9 $\pm$ 4.4 & 35.7 $\pm$ 4.4 & 38.2 $\pm$ 6.6 & 2.6 $\pm$ 0.4 \\
 & FRST-style detector~\cite{sundaresan}\textsuperscript{R} & 67.3 $\pm$ 10.5 & 66.8 $\pm$ 10.3 & 63.3 $\pm$ 9.4 & 0.9 $\pm$ 0.3 & \underline{77.2 $\pm$ 0.6} & 25.4 $\pm$ 4.1 & 35.3 $\pm$ 4.8 & 5.8 $\pm$ 0.8 \\
Our Method & CenSynCMB & \textbf{75.3 $\pm$ 8.9} & \textbf{79.3 $\pm$ 8.7}\textsuperscript{*} & \textbf{74.3 $\pm$ 8.8}\textsuperscript{*} & \textbf{0.8 $\pm$ 0.5} & \textbf{88.5 $\pm$ 6.9}\textsuperscript{*} & \textbf{55.9 $\pm$ 7.8} & \textbf{65.0 $\pm$ 6.9}\textsuperscript{*} & \underline{2.4 $\pm$ 0.6}\textsuperscript{*} \\
\hline
Contextual reference & MixMicrobleedNet~\cite{mixmicrobleednet}\textsuperscript{L} & 64.5 $\pm$ 11.1 & 70.5 $\pm$ 11.0 & 66.0 $\pm$ 11.0 & 1.0 $\pm$ 0.5 & 3.2 $\pm$ 3.7 & 7.0 $\pm$ 6.1 & 4.0 $\pm$ 3.9 & 1.0 $\pm$ 0.5 \\
 & SHIVA-CMB v2~\cite{shivacmb}\textsuperscript{L} & 47.6 $\pm$ 12.6 & 54.8 $\pm$ 13.3 & 48.9 $\pm$ 12.1 & 1.0 $\pm$ 0.5 & 79.2 $\pm$ 6.9 & 69.0 $\pm$ 7.2 & 70.5 $\pm$ 8.3 & 1.0 $\pm$ 0.5 \\
\hline
\end{tabular}}
\vspace{2pt}
\begin{minipage}{0.99\textwidth}
\footnotesize
Values are macro means $\pm$ 95\% bootstrap half-widths. Arrows indicate the preferred direction. Bold indicates the best result and underline the second-best result among leakage-free local comparators and CenSynCMB; contextual pretrained references are shown below the ranked block and excluded from ranking. Asterisks in the Our Method row indicate $p<0.05$ against the best local comparator for that metric. AIBL local comparison rows report the mean over five fold checkpoints. \textsuperscript{R}Reimplemented and retrained using our cross-validation splits. \textsuperscript{L}Public pretrained reference outside the fold-wise protocol.
\end{minipage}
\end{table*}

On VALDO, CenSynCMB achieved the strongest local-comparison performance for all four lesion-level metrics. The gains over the best non-proposed local comparator were statistically significant for precision ($p=0.0226$) and F1 ($p=0.0204$), while the recall and FP/subject differences did not reach significance. This pattern is useful for interpreting the design goal: CenSynCMB did not simply trade precision for sensitivity on the development cohort, but improved the object-level balance captured by F1.

External AIBL testing exposed a different operating point. CenSynCMB produced the highest recall and F1 among local methods, with significant gains over the best local comparators ($p=0.0058$ and $p=0.0016$), although DynUNet retained the lowest FP/subject. This pattern highlights the external recall-precision shift: CenSynCMB improved lesion recovery on SWI, but the higher sensitivity came with a larger FP burden.

\subsubsection{Patient Level Analysis}

Table~\ref{tab:patient-burden-analysis} evaluates whether lesion-level detections also supported patient-level CMB burden estimation. These results were less uniformly favourable than the lesion-level comparison, showing that better candidate recovery did not automatically translate into better count calibration.

\begin{table*}[t]
\centering
\caption{Patient-level burden regression.}
\label{tab:patient-burden-analysis}
\scriptsize
\setlength{\tabcolsep}{2pt}
\renewcommand{\arraystretch}{1.12}
\resizebox{\textwidth}{!}{%
\begin{tabular}{llccc@{\hspace{12pt}}ccc}
\hline
 &  & \multicolumn{3}{c}{VALDO (n=72)} & \multicolumn{3}{c}{AIBL (n=370)} \\
\cline{3-5}\cline{6-8}
Group & Method & MAE $\downarrow$ & RMSE $\downarrow$ & Pearson $r$ $\uparrow$ & MAE $\downarrow$ & RMSE $\downarrow$ & Pearson $r$ $\uparrow$ \\
\hline
Baseline & DynUNet~\cite{nnunet} & 2.07 $\pm$ 1.36 & 6.68 $\pm$ 4.67 & 0.67 $\pm$ 0.17 & \textbf{0.93 $\pm$ 0.14} & \textbf{1.84 $\pm$ 0.45} & \textbf{0.53 $\pm$ 0.25} \\
 & Attention U-Net~\cite{attentionunet} & 2.28 $\pm$ 1.51 & 6.91 $\pm$ 4.95 & 0.62 $\pm$ 0.23 & 6.68 $\pm$ 0.40 & 8.10 $\pm$ 0.73 & 0.41 $\pm$ 0.18 \\
 & SwinUNETR~\cite{swinunetr} & 3.40 $\pm$ 1.87 & 8.75 $\pm$ 4.82 & 0.39 $\pm$ 0.25 & 10.26 $\pm$ 0.51 & 11.82 $\pm$ 0.74 & 0.34 $\pm$ 0.15 \\
 & VISTA3D~\cite{vista3d} & 3.11 $\pm$ 1.82 & 8.28 $\pm$ 4.81 & 0.48 $\pm$ 0.40 & 7.43 $\pm$ 0.49 & 9.19 $\pm$ 1.03 & 0.41 $\pm$ 0.21 \\
\hline
VALDO methods & Zihao VALDO~\cite{cmbseg}\textsuperscript{R} & 2.92 $\pm$ 1.72 & 8.17 $\pm$ 4.79 & 0.45 $\pm$ 0.29 & 6.82 $\pm$ 0.42 & 8.26 $\pm$ 0.77 & 0.41 $\pm$ 0.19 \\
\hline
Current methods & Al-Masni-style detector~\cite{almasni}\textsuperscript{R} & 3.94 $\pm$ 1.76 & 8.65 $\pm$ 4.45 & 0.32 $\pm$ 0.31 & 6.09 $\pm$ 0.32 & 7.16 $\pm$ 0.54 & 0.41 $\pm$ 0.13 \\
 & Kim-style detector~\cite{kim2025}\textsuperscript{R} & 2.17 $\pm$ 1.33 & 6.32 $\pm$ 4.13 & \underline{0.78 $\pm$ 0.20} & 2.26 $\pm$ 0.17 & \underline{3.13 $\pm$ 0.47} & 0.46 $\pm$ 0.18 \\
 & FRST-style detector~\cite{sundaresan}\textsuperscript{R} & \underline{2.00 $\pm$ 1.23} & \underline{5.93 $\pm$ 4.09} & \textbf{0.84 $\pm$ 0.16} & 4.68 $\pm$ 0.37 & 6.11 $\pm$ 0.87 & \underline{0.47 $\pm$ 0.19} \\
Our Method & CenSynCMB & \textbf{1.81 $\pm$ 1.22} & \textbf{5.80 $\pm$ 4.06} & 0.77 $\pm$ 0.12 & \underline{1.87 $\pm$ 0.37}\textsuperscript{*} & 4.12 $\pm$ 1.53 & 0.43 $\pm$ 0.22 \\
\hline
Contextual reference & MixMicrobleedNet~\cite{mixmicrobleednet}\textsuperscript{L} & 1.10 $\pm$ 0.81 & 3.81 $\pm$ 2.66 & 0.94 $\pm$ 0.05 & 0.42 $\pm$ 0.16 & 1.56 $\pm$ 0.67 & 0.05 $\pm$ 0.11 \\
 & SHIVA-CMB v2~\cite{shivacmb}\textsuperscript{L} & 1.96 $\pm$ 1.53 & 7.44 $\pm$ 5.63 & 0.67 $\pm$ 0.21 & 0.87 $\pm$ 0.25 & 2.75 $\pm$ 1.23 & 0.51 $\pm$ 0.26 \\
\hline
\end{tabular}}
\vspace{2pt}
\begin{minipage}{0.99\textwidth}
\footnotesize
Burden regression used predicted versus ground-truth lesion counts. Values are means $\pm$ 95\% bootstrap half-widths. Arrows indicate the preferred direction. Bold indicates the best result and underline the second-best result among leakage-free local comparators and CenSynCMB; contextual pretrained references are shown below the ranked block and excluded from ranking. Asterisks in the Our Method row indicate $p<0.05$ against the best local comparator for that metric. AIBL local comparison rows report the mean over five fold checkpoints. \textsuperscript{R}Reimplemented and retrained using our cross-validation splits. \textsuperscript{L}Public pretrained reference outside the fold-wise protocol.
\end{minipage}
\end{table*}

On VALDO, CenSynCMB gave the lowest MAE and RMSE among local methods, but the MAE difference from the strongest local comparator was not significant. Pearson correlation remained close to the Kim-style detector but below the FRST-style detector, indicating that lower absolute count error did not fully determine count association. On AIBL, the higher-recall detector over-estimated burden more often, producing a significantly higher MAE than DynUNet and a larger RMSE than DynUNet and the Kim-style detector. We therefore treat patient-level burden as a calibration limitation of CenSynCMB rather than as a solved consequence of lesion-level sensitivity.

\subsection{Ablation Studies}

\subsubsection{Centre-Guided Detector}
Fig.~\ref{fig:grad-cam} illustrates how the probability and centre-map heads respond around CMB candidates, while Table~\ref{tab:stage1-ablation} isolates their quantitative effect on VALDO Task~2 together with FN-driven reweighting. The centre-map head alone increased F1 from 65.8\% to 70.3\%, and the full centre-guided detector reached 71.7\%. The paired test for this final-versus-backbone F1 difference was close to, but did not meet, the conventional significance threshold ($p=0.0986$), so we interpret the ablation as directional support for the detector design rather than as definitive statistical separation.

\begin{table}[!htbp]
\centering
\caption{centre-guided detector ablation.}
\label{tab:stage1-ablation}
\scriptsize
\setlength{\tabcolsep}{2pt}
\renewcommand{\arraystretch}{1.12}
\resizebox{\columnwidth}{!}{%
\begin{tabular}{p{0.42\columnwidth}cccc}
\hline
Configuration & Recall $\uparrow$ (\%) & Precision $\uparrow$ (\%) & F1 $\uparrow$ (\%) & FP/subject $\downarrow$ \\
\hline
Backbone & 69.7 $\pm$ 10.3 & 68.6 $\pm$ 10.3 & 65.8 $\pm$ 10.2 & 1.4 $\pm$ 0.9 \\
Backbone + FN-RW & 71.5 $\pm$ 10.4 & 69.2 $\pm$ 10.7 & 67.6 $\pm$ 9.5 & 1.9 $\pm$ 1.2 \\
Backbone + CM & 76.4 $\pm$ 9.6 & 71.6 $\pm$ 9.4 & 70.3 $\pm$ 9.1 & \textbf{1.3 $\pm$ 0.8} \\
Backbone + CM + FN-RW & \textbf{76.9 $\pm$ 9.3} & \textbf{74.6 $\pm$ 9.6} & \textbf{71.7 $\pm$ 9.2} & 1.4 $\pm$ 1.0 \\
\hline
\end{tabular}}
\vspace{2pt}
\begin{minipage}{0.98\columnwidth}
\footnotesize
Backbone denotes the Attention U-Net detector; CM and FN-RW denote centre-map supervision and false-negative-driven reweighting, respectively. Values are macro means $\pm$ 95\% bootstrap half-widths. Arrows indicate the preferred direction and bold marks the best result in each column.
\end{minipage}
\end{table}

\subsubsection{Physics-Guided Synthesis}
\label{subsec:data-synthesis}
The visual examples in Fig.~\ref{fig:synthetic-examples} illustrate the positive CMB and mimic phenotypes introduced by the renderer. Table~\ref{tab:data-synthesis-ablation} then evaluates how these synthetic positives and hard negatives change the detector when added to the final real-data configuration. Synthetic positive CMBs mainly improved precision and F1 relative to real data alone, while the addition of hard negatives produced the highest F1 and the lowest FP/subject. None of the paired synthesis ablation tests reached $p<0.05$, but the direction of change matches the intended role of the mimic labels: they reduce false positives without collapsing lesion recovery.

\begin{table}[t]
\centering
\caption{physics-guided synthesis ablation.}
\label{tab:data-synthesis-ablation}
\scriptsize
\setlength{\tabcolsep}{2pt}
\renewcommand{\arraystretch}{1.12}
\resizebox{\columnwidth}{!}{%
\begin{tabular}{p{0.42\columnwidth}cccc}
\hline
Training Data & Recall $\uparrow$ (\%) & Precision $\uparrow$ (\%) & F1 $\uparrow$ (\%) & FP/subject $\downarrow$ \\
\hline
Real Data Only & \textbf{76.9 $\pm$ 9.3} & 74.6 $\pm$ 9.6 & 71.7 $\pm$ 9.2 & 1.4 $\pm$ 1.0 \\
Real Data + Syn. CMB & 73.6 $\pm$ 10.4 & 77.0 $\pm$ 10.1 & 73.1 $\pm$ 10.3 & 1.0 $\pm$ 0.6 \\
Real Data + Syn. CMB + Mimics & 75.3 $\pm$ 8.9 & \textbf{79.3 $\pm$ 8.7} & \textbf{74.3 $\pm$ 8.8} & \textbf{0.8 $\pm$ 0.5} \\
\hline
\end{tabular}}
\vspace{2pt}
\begin{minipage}{0.98\columnwidth}
\footnotesize
All rows use the Backbone + CM + FN-RW detector from Table~\ref{tab:stage1-ablation}. Syn. CMB and Mimics denote synthetic positive CMBs and synthetic hard negatives added to the real training data. Values are macro means $\pm$ 95\% bootstrap half-widths. Arrows indicate the preferred direction and bold marks the best result in each column.
\end{minipage}
\end{table}

\section{Discussion}

\subsection{Cohort-Scale Detection Use}
CenSynCMB is best viewed as a detection-oriented component for scalable CMB cohort analysis. Its strongest evidence was lesion-level F1 on both VALDO and AIBL, showing that the detector improved the balance between recall and precision under centroid-based evaluation. This is the level at which errors are most consequential for downstream cohort work: a missed CMB cannot be recovered by later review, whereas an over-called candidate can still be rejected by threshold adjustment, visual inspection, or cohort-specific calibration.

This framing matches our intended use case. Large, unlabelled cohort studies cannot usually support exhaustive expert annotation of every susceptibility-sensitive scan, but they can benefit from a model that extracts CMB candidates, prioritises uncertain scans, and supports semi-automated burden estimation. The AIBL results suggest that this goal is plausible across a GRE-to-SWI shift, but they also show that candidate recovery is not the same as calibrated burden estimation. For population-level analysis, the detection output should therefore be treated as a high-throughput candidate layer rather than an unqualified clinical count.

\subsection{Centre Guidance and Mimic Synthesis}
The two main design choices address complementary failure modes. Centre-map supervision brings training closer to the centroid-localisation target used in VALDO-style evaluation, and the Grad-CAM examples and the detector ablation are both consistent with improved centroid localisation. This does not make the architecture novel by itself; rather, it makes the training objective better aligned with how compact CMB candidates are ultimately judged.

Mimic-aware synthesis focuses on false positives from clinically plausible hypointense structures. Synthetic CMBs increase exposure to rare compact lesions, while calcification-like and vessel-like hard negatives expose the model to common hypointense mimics that are difficult to sample exhaustively from VALDO alone. The synthesis ablation moved precision, F1, and FP/subject in the intended direction, although the evidence remains trend-level rather than statistically definitive. The value of the synthesis module is therefore not only additional data volume, but controlled exposure to labelled mimic phenotypes.

\subsection{Limitations and Future Work}
The current system still has important limits. VALDO is small and CMB-positive enriched, AIBL is SWI-only, and patient-level burden estimates remained sensitive to false-positive accumulation. For ARIA-H translation, CenSynCMB covers CMBs but not cortical superficial siderosis, so it cannot by itself quantify the full haemorrhagic ARIA-H spectrum. Recent multicentre ARIA-H work similarly emphasises that automated CMB tools must support both lesion detection and reliable burden quantification~\cite{low2026}. Future use in large unlabelled cohorts should therefore include cohort-specific threshold calibration, uncertainty-aware candidate review, or active learning from a small expert-checked subset. These steps are not peripheral deployment details; they are necessary for translating a sensitive candidate detector into reliable cohort-level burden estimates.

The renderer is also a susceptibility-inspired magnitude approximation rather than a full GRE, SWI, phase, or QSM simulator. In particular, magnitude-only static dephasing cannot distinguish calcification from blood products through susceptibility sign, which is why clinical reading often relies on phase or QSM for this distinction. Incorporating sequence metadata, phase information, or QSM-derived priors could improve the realism of both positive lesions and hard negatives.

\section{Conclusion}
We presented CenSynCMB, a centre-guided and mimic-aware framework for cerebral microbleed detection. By combining an Attention U-Net detector with auxiliary centre-map supervision, false-negative-driven reweighting, and fold-wise physics-guided synthesis, CenSynCMB improved lesion-level F1 on VALDO and on the external AIBL SWI cohort. The results also define the present boundary of the system: it is a strong detection-oriented component, but patient-level burden calibration and false-positive control remain necessary before application to large unlabelled cohort studies. These findings support a practical route toward scalable CMB candidate extraction, semi-automated review, and future population-level imaging analysis.

\section*{Acknowledgment}
We acknowledge support from the UCL EPSRC CDT in i4health (EP/S021930/1) and the British Heart Foundation through the UCL BHF Research Accelerator Award (AA/18/6/34223), and use of the Isambard platform.

\IEEEtriggeratref{15}

\end{document}